\documentclass[journal]{IEEEtran}

\ifCLASSINFOpdf
\else
   \usepackage[dvips]{graphicx}
\fi
\usepackage{url}

\usepackage{amsmath,graphicx,amssymb}

\begin{document}

\title{Composed Vision-Language Retrieval for Skin Cancer Case Search via Joint Alignment of Global and Local Representations}

\author{Yuheng Wang, Yuji Lin, Jiayue Cai, Z. Jane Wang, Tim K. Lee}


\markboth{Wang \MakeLowercase{\textit{et al.}} }
{Shell \MakeLowercase{\textit{et al.}}: Bare Demo of IEEEtran.cls for IEEE Journals}
\maketitle

\begin{abstract}
Medical image retrieval aims to identify clinically relevant lesion cases to support diagnostic decision making, education, and quality control. In practice, retrieval queries often combine a reference lesion image with textual descriptors such as dermoscopic features. We study composed vision–language retrieval for skin cancer, where each query consists of an image–text pair and the database contains biopsy-confirmed, multi-class disease cases. We propose a transformer-based framework that learns hierarchical composed query representations and performs joint global–local alignment between queries and candidate images. Local alignment aggregates discriminative regions via multiple spatial attention masks, while global alignment provides holistic semantic supervision. The similarity is computed through a convex, domain-informed weighting that emphasizes clinically salient local evidence while preserving global consistency. Experiments on the public Derm7pt dataset demonstrate consistent improvements over state-of-the-art methods. The proposed framework enables efficient access to relevant medical records and supports practical clinical deployment.
\end{abstract}


\IEEEpeerreviewmaketitle

\section{Introduction}

\begin{figure}[t]
\centering
\includegraphics[width=0.85\linewidth]{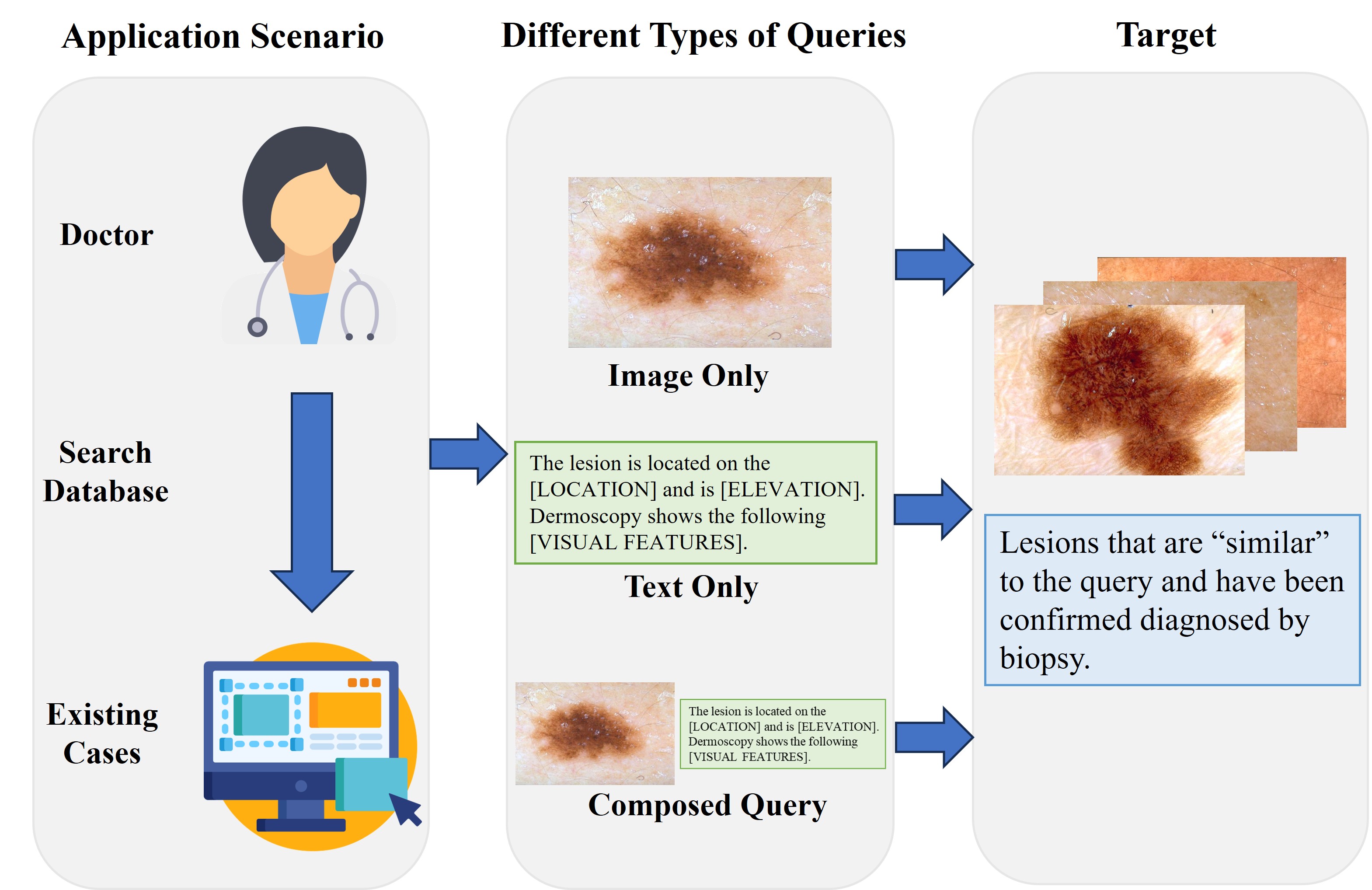}
\caption{Types of clinical skin cancer case search and query. Conventional retrieval uses image-only queries or text-only clinical descriptors separately, whereas our composed retrieval pairs a reference lesion image with its associated text to form a vision-language query. All settings aim to retrieve visually similar, biopsy-confirmed cases from an image-only database to support clinical decision making.}
\label{fig:intro}
\end{figure}

\IEEEPARstart{E}{arly} detection and diagnosis of skin cancer enable timely treatment planning, reducing patient morbidity and improving cure rates~\cite{celebi2019dermoscopy,chiou2025multimodal}. Achieving these benefits, however, depends on accurate interpretation of often subtle lesion appearances. With rapid advances in artificial intelligence, deep learning-based classification systems have reached dermatologist-level performance in multiple settings~\cite{esteva2017dermatologist,barata2023reinforcement,yan2025multimodal,wang2025neural,kurtansky2025automated}. However, a central challenge remains translating these models into routine clinical workflows. Compared with end-to-end classification, case-based retrieval offers a more clinically intuitive form of decision support by comparing similar historical cases, thereby facilitating comparative assessment, interpretability, and clinician training~\cite{tschandl2019diagnostic,zhang2022dermoscopic,ozbay2023interpretable,chen2023multi,chen2025deep}. In practice, clinicians generally formulate retrieval queries by pairing a reference lesion image with concise textual descriptions (e.g., dermoscopic patterns or checklist-based criteria)~\cite{wang2022multi}. Thus, constructing effective vision–language hybrid queries that fully exploit multimodal information is critical to enabling practical, trustworthy skin cancer decision support, as shown in Fig.~\ref{fig:intro}.

\begin{figure*}[t]
\centering
\includegraphics[width=0.9\linewidth]{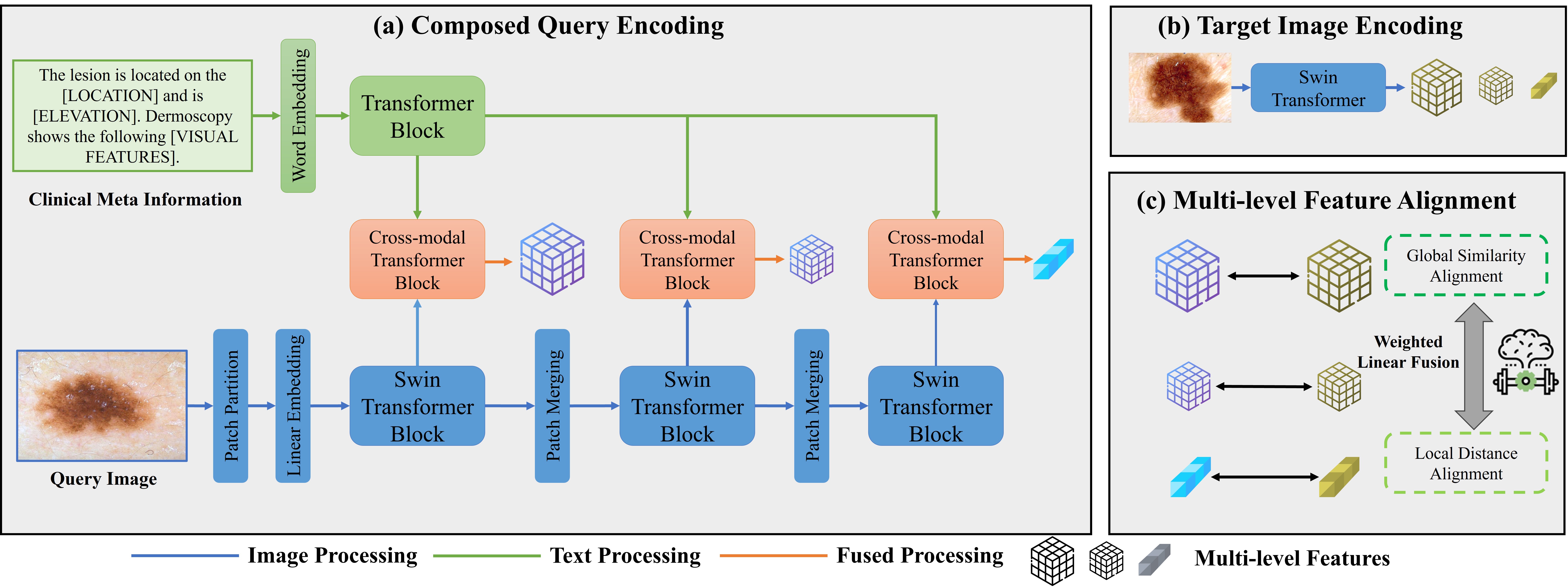}
\caption{Overall workflow of composed vision-language retrieval for skin cancer.
The query image and text are fused via cross-modal Transformers on top of a hierarchical vision backbone to form multi-level composed query representations (a), while each database image is encoded by the same backbone (b). Multi-level global and local alignment jointly compute query-target similarity for retrieval ranking (c).
}
\label{fig:workflow}
\end{figure*}

Over the past five years, transformer architectures, which are well-suited to modeling long-range dependencies in sequential data, have driven major advances in large-scale semantic prediction models~\cite{vaswani2017attention,koroteev2021bert,rasmy2021med,zhou2024pre}. Building on these developments, vision transformers have further reshaped performance benchmarks in computer vision by enabling effective global context modeling in images~\cite{liu2021swin,xu2023multi,xu2024skinformer,pandala2026optimal}. These advances provide a strong technical foundation for multimodal composed image retrieval, where visual and textual cues can be integrated into a unified representation for reliable case matching~\cite{yang2023composed,song2025comprehensive,guo2026textbridge}.

Based on the above motivation and background, we study transformer-based composed vision-language retrieval for skin cancer case search, where each query pairs a reference lesion image with accompanying clinical text to retrieve relevant cases from an image database. The key challenge is to design a clinically meaningful similarity function that captures global semantics (e.g., overall lesion morphology and color distribution) while emphasizing localized discriminative cues (e.g., streaks, irregular pigmentation, and regression-like structures). We address this by learning multi-level composed query representations via image and text fusion and matching them to target images through joint global-local similarity alignment, enabling robust case ranking for clinical assessment. In this letter, our main contributions are as follows: 

\begin{itemize}
\item We formulate skin cancer case search as a composed vision-language retrieval problem, where each query pairs a reference lesion image with its associated clinical text to retrieve target images from the existing database.

\item We adapt a hierarchical representation learning framework with joint global-local alignment for skin cancer retrieval, where learnable region masks capture discriminative local patterns and a weighted global-local similarity emphasizes clinically relevant details.

\item We conduct both quantitative and qualitative experiments on a publicly available multimodal skin lesion dataset and demonstrate state-of-the-art retrieval performance.
\end{itemize}

\section{Method}

Let the image database be $\mathcal{D}=\{(I_n, y_n)\}_{n=1}^{N}$, where $I_n$ denotes a lesion image and $y_n$ is its diagnostic label. 
A retrieval query is defined as $q=(I_q, T_{\tau}, y_q)$, consisting of a \emph{query} image $I_q$ and a textual description $T_{\tau}$ (e.g., checklist-based criteria) associated with the query label $y_q$. 
We define label-level positives $\mathcal{P}$ as
\begin{equation}
\mathcal{P}(q)=\{n \mid y_n = y_q\}
\end{equation}

Given $q$, our goal is to learn a similarity function $S(q, I_n)$ that ranks all database images $I_n$ and returns the top-$K$ most relevant cases with the same diagnosis label. 
This formulation reflects the clinical workflow where the image provides morphological context while the text specifies discriminative diagnostic cues.

Fig.~\ref{fig:workflow} illustrates the overall workflow of our framework. Given a composed query $q=(I_q, T_{\tau}, y_q)$ and a candidate target image $I_t$, our model computes a multi-level similarity by integrating (i) hierarchical visual representations, (ii) cross-modal composition between $I_q$ and $T_{\tau}$, and (iii) joint global/local alignment between the composed query and $I_t$. 
Concretely, we first extract multi-level feature maps from both reference and target images using a hierarchical vision backbone. 
We then fuse the reference-image features with text token embeddings via a cross-modal Transformer to obtain a composed query representation at each feature level. 
Finally, we measure query-target similarity through complementary global and local alignment terms: global alignment captures overall semantics and provides stable supervision, whereas local alignment emphasizes region-level discriminative patterns that are critical for clinical diagnosis.

\subsection{Hierarchical visual encoding}
As illustrated in Fig.~\ref{fig:workflow}(a), we use a hierarchical vision backbone built on the Swin Transformer~\cite{liu2021swin} to extract multi-level visual features, following the practice in prior composed retrieval frameworks~\cite{xu2023multi}. The extracted multi-level visual features $f_{\text{vis}}$ from an image $I$:
\begin{equation}
\{X^{L}(I), X^{M}(I), X^{H}(I)\} = f_{\text{vis}}(I)
\end{equation}
where $X^{L}\in\mathbb{R}^{h_L\times w_L\times d_L}$, $X^{M}\in\mathbb{R}^{h_M\times w_M\times d_M}$, and $X^{H}\in\mathbb{R}^{h_H\times w_H\times d_H}$ denote low-, middle-, and high-level feature maps, respectively. 
These hierarchical representations preserve both fine-grained appearance details and higher-level semantic context, which are jointly necessary for robust lesion matching.
For the query image $I_q$ and a target database image $I_t$, we compute
\begin{equation}
X_q^{i} = X^{i}(I_q),\quad X_t^{i} = X^{i}(I_t),\quad i\in\{L,M,H\}
\end{equation}

\subsection{Text encoding and cross-modal composition}
Given the textual description $T_{\tau}$, we employ BERT~\cite{liu2019roberta} as the language encoder to obtain token embeddings, following the text encoding setup in~\cite{wang2025integrating}:
\begin{equation}
Z_{\tau} = f_{\text{text}}(T_{\tau})\in\mathbb{R}^{n\times d_T}
\end{equation}
where $n$ is the number of tokens and $d_T$ is the token embedding dimension. 
To construct a composed query that reflects both the reference appearance and the specified textual attributes, we employ a cross-modal Transformer that injects textual information into the reference visual features. 
For each feature level $i\in\{L,M,H\}$, we compute
\begin{equation}
X_{q\tau}^{i} = \mathrm{CrossTransformer}(X_{q}^{i}, Z_{\tau})
\end{equation}
where $X_{q\tau}^{i}\in\mathbb{R}^{h_i\times w_i\times d_i}$ denotes the composed query feature map at level $i$, expressed in the same visual feature space as $X_t^{i}$ to enable direct query-target alignment.


\subsection{Joint alignment of multi-level representations}
We quantify the similarity between the composed query $\{X_{q\tau}^{i}\}$ and the target image features $\{X_{t}^{i}\}$ via two complementary terms. 
Global alignment compares pooled representations to capture overall lesion-level semantics (e.g., morphology and color distribution). 
Local alignment, in contrast, focuses on discriminative subregions (e.g., streaks, irregular pigmentation, regression-like patterns) by learning multiple region masks and aggregating region descriptors. 
Combining these two views yields a similarity function that is both semantically coherent and locally sensitive.

Firstly, we model local discriminative patterns by learning $k$ region masks and aggregating region descriptors. 
For each feature level $i\in\{L,M,H\}$ and region index $j\in\{1,\ldots,k\}$, a region descriptor is computed as
\begin{equation}
E_{ij}=\rho\!\left(X^{i}\odot \alpha_{j}(X^{i})\right)
\label{eq:local_region}
\end{equation}
where $\alpha_j(\cdot)\in\mathbb{R}^{h_i\times w_i\times 1}$ produces a spatial attention mask, $\odot$ denotes Hadamard product, and $\rho(\cdot)$ is a pooling operator yielding a $d_i$-dimensional vector. 
Applying Eq.~\eqref{eq:local_region} to $X_{q\tau}^{i}$ and $X_{t}^{i}$ yields region sets $E_{q\tau}^{i}\in\mathbb{R}^{k\times d_i}$ and $E_{t}^{i}\in\mathbb{R}^{k\times d_i}$. 
We then compute the local similarity by aggregating the $k$ region descriptors and measuring cosine similarity:
\begin{equation}
S_{\text{local}}=\sum_{i\in\{L,M,H\}} s\!\left(\rho(E_{q\tau}^{i}),\,\rho(E_{t}^{i})\right)
\label{eq:s_local}
\end{equation}
where $s(\cdot,\cdot)$ denotes cosine similarity and $\rho(\cdot)$ aggregates region descriptors (e.g., average over $k$). This formulation encourages the model to align diagnostically informative patterns without requiring explicit lesion-level annotations.

For the global alignment, to capture overall semantics and provide stable supervision complementary to local matching, we compute a global similarity by pooling each feature map and applying cosine similarity: 

\begin{equation} 
S_{\text{global}}=\sum_{i\in\{L,M,H\}} s\!\left(\rho(X_{q\tau}^{i}),\,\rho(X_{t}^{i})\right) 
\label{eq:d_global}  
\end{equation} 

Global alignment emphasizes holistic consistency between the composed query and the target image, and empirically helps stabilize the learning of local attention masks by preventing degenerate region selections. We then combine the local and global similarities via a weighted aggregation to obtain the final query-target similarity score:
\begin{equation}
S = \beta\, S_{\text{local}} + (1-\beta)\, S_{\text{global}}
\label{eq:s_total}
\end{equation}
where $\beta\in[0,1]$ governs the trade-off between local and global signals, and a larger $S$ indicates a stronger query--target match. 
In skin cancer case search, diagnostically salient cues are frequently confined to specific lesion subregions (e.g., streaks, irregular pigmentation, and regression-like patterns); accordingly, we choose a $\beta$ that prioritizes $S_{\text{local}}$ while using $S_{\text{global}}$ as a holistic semantic constraint. 

This design departs from prior composed/image retrieval methods that predominantly rely on global embedding similarity \cite{tschandl2019diagnostic,chen2025deep,wang2022multi} or perform global--local aggregation without accounting for domain-dependent cue reliability \cite{xu2023multi,weng2025unsupervised}. By embedding this clinical reliability prior into a simple convex fusion, our similarity better aligns with medical decision patterns without sacrificing global consistency.

\begin{table}[t]
\centering
\caption{Accuracy@K overall performance across 5-fold cross-validation (Mean $\pm$ Std, in \%).}
\label{tab:retrieval_acc}
\renewcommand{\arraystretch}{1.2}

\resizebox{\columnwidth}{!}{%
\begin{tabular}{lccc}
\hline
\textbf{Method} 
& \textbf{Acc@1} 
& \textbf{Acc@2} 
& \textbf{Acc@4} \\
\hline
ResNet50-CosSim~\cite{tschandl2019diagnostic} 
& 77.6 $\pm$ 3.5 
& 80.9 $\pm$ 2.6 
& 83.2 $\pm$ 1.9 \\

SNF-DCA~\cite{wang2022multi}
& 77.8 $\pm$ 2.3
& \textbf{82.8 $\pm$ 1.4}
& 83.3 $\pm$ 2.5 \\

MaskRCNN-Fusion~\cite{iqbal2023fusion}
& 62.5 $\pm$ 1.2 
& 77.3 $\pm$ 3.1 
& \textbf{87.3 $\pm$ 2.5} \\

DAHNET~\cite{jiang2024global}
& 74.4 $\pm$ 2.1  
& 74.9 $\pm$ 1.8 
& 75.6 $\pm$ 2.7 \\
DEML~\cite{weng2025unsupervised}
& 76.3 $\pm$ 1.3 
& 80.2 $\pm$ 1.5  
& 84.3 $\pm$ 2.2 \\
WalshHash~\cite{chen2025deep}
& 77.5 $\pm$ 2.0  
& 81.6 $\pm$ 1.6
& 84.9 $\pm$ 1.2 \\

Proposed
& \textbf{79.3 $\pm$ 2.9}
& 82.4 $\pm$ 3.6
& \textbf{87.3 $\pm$ 3.2} \\

\hline
\end{tabular}%
}
\end{table}

\begin{table}[t]
\centering
\caption{Mean Average Precision (mAP) for each fold and overall performance across 5-fold cross-validation (in \%).}
\label{tab:retrieval_map}
\renewcommand{\arraystretch}{1.2}

\resizebox{\columnwidth}{!}{%
\begin{tabular}{lccccc|c}
\hline
\textbf{Method}
& \textbf{Fold 1}
& \textbf{Fold 2}
& \textbf{Fold 3}
& \textbf{Fold 4}
& \textbf{Fold 5}
& \textbf{Mean $\pm$ Std} \\
\hline

ResNet50-CosSim~\cite{tschandl2019diagnostic}
& 78.5 & 81.7 & \textbf{84.0} & 77.2 & 80.6
& 80.4 $\pm$ 2.7 \\

SNF-DCA~\cite{wang2022multi}
& 79.4 & 81.6 & 80.4 & 80.1 & 78.4
& 80.0 $\pm$ 1.2 \\

MaskRCNN-Fusion~\cite{iqbal2023fusion}
& 61.3 & 57.8 & 60.4 & 61.5 & 61.2
& 60.5 $\pm$ 1.5 \\

DAHNET~\cite{jiang2024global}
& 79.1 & 80.5 & 81.6 & \textbf{80.2} & \textbf{81.5}
& 80.6 $\pm$ 1.0 \\

DEML~\cite{weng2025unsupervised}
& 80.1& 81.4 & 82.2 & 78.4 & 79.8
& 80.4 $\pm$ 1.5 \\
WalshHash~\cite{chen2025deep}
& 80.6 & 82.3 & 80.4 & 79.9 & 80.6
& 80.8 $\pm$ 0.9\\

Proposed
& \textbf{83.8} & \textbf{84.5} & 82.4 & 78.6 & 79.3
& \textbf{81.7 $\pm$ 2.7} \\
\hline
\end{tabular}%
}
\end{table}


\begin{figure*}[t]
\centering
\includegraphics[width=0.9\linewidth]{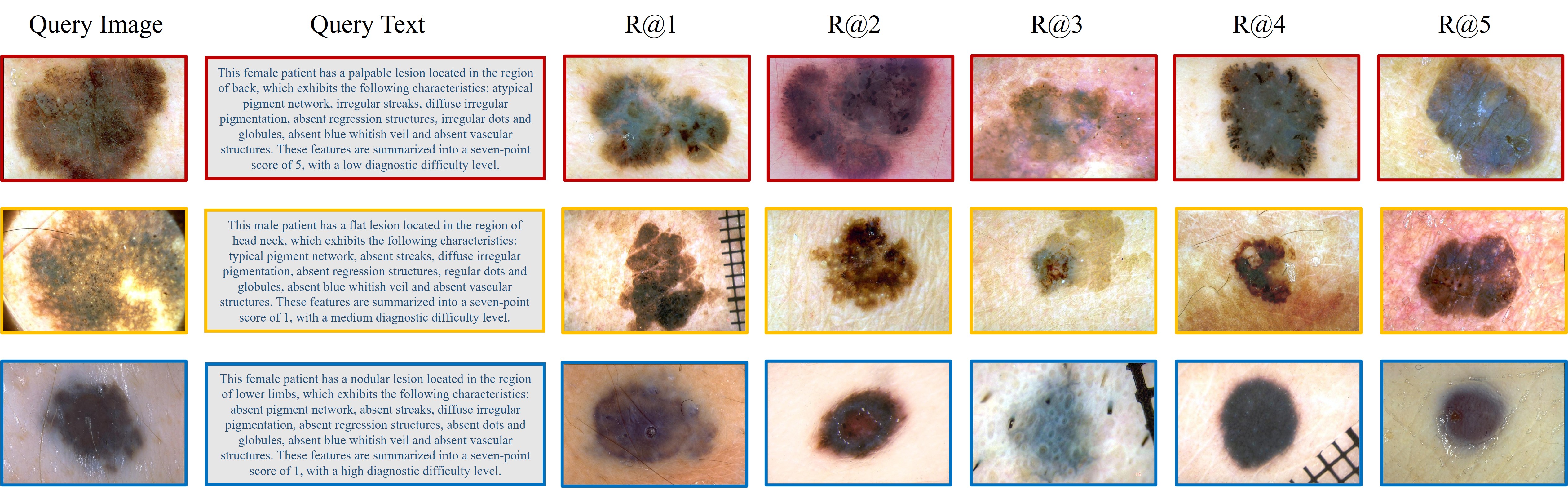}
\caption{Qualitative composed retrieval examples on Derm7pt. Each row shows one query (left: query image; middle: associated clinical text) and the top-5 retrieved images (right, R@1--R@5). Rows 1--3 correspond to mel (red borders), bkl (yellow borders), and nevus (blue borders), respectively.}

\label{fig:results}
\end{figure*}

\section{Experiments}

We selected Derm7pt, a widely used benchmark for skin cancer research, because it provides dermoscopic images alongside patient metadata and structured clinical attributes aligned with the 7-point checklist~\cite{kawahara2018seven}. Following prior retrieval-oriented studies, we adopted the standard evaluation protocol for retrieval on this dataset. To ensure sufficient class representation and reliable performance estimation, we restricted analysis to diagnostic categories with at least 50 samples~\cite{tschandl2019diagnostic}. Under this criterion, Derm7pt yields 888 eligible images from the original 1,011, spanning three categories: melanoma (mel), nevus, and benign keratosis-like lesions (bkl). For the original settings of the dataset, each image has a maximum resolution of 768 × 512 pixels. Since our method uses only dermoscopy images and meta information, we ensured fair comparisons with state-of-the-art approaches by adopting the same input setting. Specifically, for methods requiring image and metadata inputs (e.g., SNF-DCA), we consistently used dermoscopy images together with the associated meta information, and excluded clinical photographs from the dataset for all methods. In the training phase, we adopted cross-validation to obtain reliable performance estimates under the limited dataset size. We set the number of epochs to 100 and employed the Adam optimizer with a learning rate of $10^{-4}$ and a weight decay of $10^{-5}$. Early stopping was applied with a patience of 30 epochs. The weighting coefficient $\beta$ in Eq.~\eqref{eq:s_total} was set as 0.6 based on preliminary experiments. For the evaluation metrics, we assess retrieval performance using mean Average Precision (mAP) and Accuracy@\(K\) following standard practice in prior works~\cite{wang2022multi,jiang2024global}.

\section{Results and Discussion}

\subsection{Quantitative results}

We compare our method with ResNet50-CosSim~\cite{tschandl2019diagnostic} as the benchmark baseline on this dataset, SNF-DCA~\cite{wang2022multi} as a representative multimodal skin cancer retrieval method, DEML~\cite{weng2025unsupervised} and MaskRCNN-Fusion~\cite{iqbal2023fusion} as recent skin-oriented retrieval approaches, and DAHNET~\cite{jiang2024global} and WalshHash~\cite{chen2025deep} as strong recent retrieval baselines. As shown in Table~\ref{tab:retrieval_acc}, our method achieves the best overall Accuracy@$\{1,2,4\}$ with an average of 83.0\%. The improvement is most meaningful at the top of the ranked list, where our Accuracy@1 reaches 79.3\%, higher than SNF-DCA at 77.8\%, ResNet50-CosSim at 77.6\%, and all other competing methods. This result is particularly important for clinical retrieval, since the first returned case often carries the greatest practical value during visual assessment and decision support. The advantage remains stable at broader cutoffs. Our method achieves 82.4\% at Accuracy@2 and 87.3\% at Accuracy@4, matching the best result while preserving the strongest early precision. In contrast, MaskRCNN-Fusion also reaches 87.3\% at Accuracy@4 but drops substantially to 62.5\% at Accuracy@1, suggesting that segmentation-driven fusion can retrieve relevant cases within a broader candidate set yet is less effective at placing the most relevant case at the top. This difference indicates that our method improves not only recall within the top retrieved samples, but also the reliability of the highest-ranked match. Table~\ref{tab:retrieval_map} shows a consistent trend in ranking quality. Our method obtains the highest mean mAP of 81.7\%, outperforming WalshHash at 80.8\%, DAHNET and ResNet50-CosSim at 80.4\%, and SNF-DCA at 80.0\%. Although the margin in mAP is moderate, the gain is consistent with the clear improvement in Accuracy@1, indicating that the proposed framework produces a better ordered ranked list rather than benefiting from a single cutoff only. Overall, these results support that combining structured clinical descriptors with localized visual evidence leads to more precise early retrieval and stronger overall ranking quality, which is well aligned with the demands of real-world skin cancer case search.

\subsection{Qualitative analysis}
Fig.~\ref{fig:results} presents representative composed retrieval results on Derm7pt. For each query, the model integrates a reference lesion image with its associated clinical text and returns the top-5 most similar cases from the image database. Overall, the retrieved cases are visually and clinically consistent with the queries: for the melanoma example (Row~1, red), top-ranked results preserve key malignant cues such as heterogeneous pigmentation and irregular structures; for the bkl example (Row~2, yellow), retrieved lesions exhibit characteristic keratotic appearance and coarse texture patterns; and for the nevus example (Row~3, blue), the returned cases show more regular and homogeneous pigmentation with smoother boundaries. Notably, the top-ranked predictions in each row remain coherent across R@1--R@5, suggesting that the proposed local-emphasized similarity effectively captures discriminative substructures while global similarity maintains overall morphological consistency. These observations align with the quantitative gains in early-rank metrics, supporting that the framework retrieves clinically relevant analogues that are suitable for comparative assessment in decision support.

\section{Conclusion}
In this work, we investigate composed image retrieval for skin cancer case search, where each query pairs an image with clinical text to retrieve relevant cases from an image database of biopsy- or expert-confirmed diagnoses. Our approach learns hierarchical composed representations and ranks targets using a weighted global-local similarity that balances holistic semantics with discriminative local cues. By returning clinically comparable, biopsy-confirmed cases, the framework provides intuitive decision support and enables efficient access to evidence.

\bibliographystyle{IEEEbib}
\bibliography{strings,refs}

\end{document}